\RequirePackage{tikz}
\documentclass{IEEEoj}
\usepackage{cite}
\usepackage{amsmath,amssymb,amsfonts}
\usepackage{algorithmic}
\usepackage{graphicx,color}
\usepackage{textcomp}
\def\BibTeX{{\rm B\kern-.05em{\sc i\kern-.025em b}\kern-.08em
    T\kern-.1667em\lower.7ex\hbox{E}\kern-.125emX}}
\AtBeginDocument{\definecolor{ojcolor}{cmyk}{0.93,0.59,0.15,0.02}}
\def\OJlogo{\vspace{-14pt}\includegraphics[height=28pt]{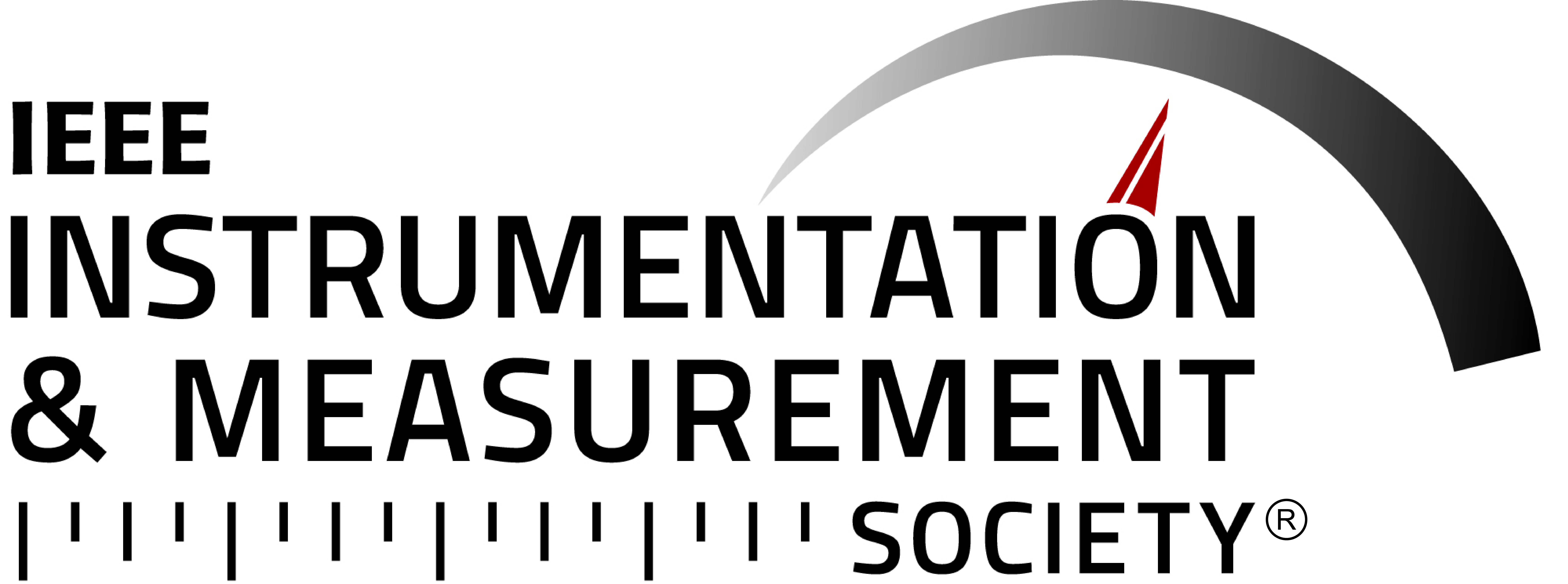}}

\usetikzlibrary{external}
\usepackage{tabularx}
\usepackage{booktabs}
\usepackage{multirow}
\usepackage{svg}
\usepackage[caption=false]{subfig}
\usepackage{pgfplots}
\pgfplotsset{compat=1.17}
\usepackage{MnSymbol}

\usepackage{xcolor}
\newcommand\crule[1]{\textcolor{#1}{\rule{6pt}{6pt}}}
\definecolor{class1_br}{rgb}{1, 0.349, 0.392}
\definecolor{class2_br}{rgb}{0.086, 0.858, 0.576}
\definecolor{class3_br}{rgb}{0.937, 0.917, 0.352}
\definecolor{class4_br}{rgb}{0, 0.486, 0.745}

\usepackage{listings}       
\definecolor{codegreen}{rgb}{0,0.6,0}
\definecolor{codegray}{rgb}{0.5,0.5,0.5}
\definecolor{codepurple}{rgb}{0.58,0,0.82}
\definecolor{backcolour}{RGB}{255, 247, 237}
\lstdefinestyle{mystyle}{
    backgroundcolor=\color{backcolour},   
    commentstyle=\color{codegreen},
    keywordstyle=\color{blue},
    numberstyle=\tiny\color{codegray},
    stringstyle=\color{codepurple},
    basicstyle=\footnotesize\ttfamily,
    breakatwhitespace=false,         
    breaklines=true,                 
    captionpos=b,                    
    keepspaces=true,                 
    numbers=left,
    frame=bt,
    numbersep=5pt,                  
    showspaces=false,                
    showstringspaces=false,
    showtabs=false,                  
    tabsize=2
}
 
\usepackage{bm}
\newcommand{\matr}[1]{\bm{#1}}      

\usepackage{xcolor}

\usepackage{tikz}
\usepackage{hyperref}
\hypersetup{
	colorlinks=true,
	linkcolor=black,     
	urlcolor=blue,
	citecolor=black,
}

\definecolor{lime}{HTML}{A6CE39}
\DeclareRobustCommand{\orcidicon}{%
	\begin{tikzpicture}
	\draw[lime, fill=lime] (0,0) 
	circle [radius=0.16] 
	node[white] {{\fontfamily{qag}\selectfont \tiny ID}};
	\draw[white, fill=white] (-0.0625,0.095) 
	circle [radius=0.007];
	\end{tikzpicture}
	\hspace{-2mm}
}
\foreach \x in {A, ..., Z}{
	\expandafter\xdef\csname orcid\x\endcsname{\noexpand\href{https://orcid.org/\csname orcidauthor\x\endcsname}{\noexpand\orcidicon}}
}
\newcommand{\orcid}[1]{\href{https://orcid.org/#1}{\textcolor[HTML]{A6CE39}{\orcidicon}}}

\usepackage[acronym]{glossaries}
\newacronym{cvnn}{CVNN}{Complex-Valued Neural Network}
\newacronym{rvnn}{RVNN}{Real-Valued Neural Network}
\newacronym{cv-mlp}{CV-MLP}{Complex-Valued MultiLayer Perceptron}
\newacronym{rv-mlp}{RV-MLP}{Real-Valued MultiLayer Perceptron}
\newacronym{mlp}{MLP}{MultiLayer Perceptron}
\newacronym{cnn}{CNN}{Convolutional Neural Network}
\newacronym{cv-cnn}{CV-CNN}{Complex-Valued Convolutional Neural Network}
\newacronym{rv-cnn}{RV-CNN}{Real-Valued Convolutional Neural Network}
\newacronym{relu}{ReLU}{Rectified Linear Unit}
\newacronym{tanh}{tanh}{hyperbolic tangent}
\newacronym{sgd}{SGD}{Stochastic Gradient Descent}
\newacronym{rmsprop}{RMSprop}{Root Mean Square Propagation}
\newacronym{polsar}{PolSAR}{Polarimetric Synthetic Aperture Radar}
\newacronym{sar}{SAR}{Synthetic Aperture Radar}
\newacronym{cv-fcnn}{CV-FCNN}{Complex-Valued Fully Convolutional Neural Network}
\newacronym{rv-fcnn}{RV-FCNN}{Real-Valued Fully Convolutional Neural Network}
\newacronym{fcnn}{FCNN}{Fully Convolutional Neural Network}
\newacronym{oa}{OA}{Overall Accuracy}
\newacronym{aa}{AA}{Average Accuracy}
\newacronym{bn}{BN}{BatchNormalization}
\newacronym{polinsar}{PolInSAR}{Polarimetric and Interferometric Synthetic Aperture Radar}

\newcommand{\loss}{\mathcal{L}}
\newcommand{\activation}{\sigma}

\renewcommand{\Im}{\mathrm{Im}}
\renewcommand{\Re}{\mathrm{Re}}

\DeclareRobustCommand*{\IEEEauthorrefmark}[1]{%
  \raisebox{0pt}[0pt][0pt]{\textsuperscript{\footnotesize\ensuremath{#1}}}}

\begin{document}
\receiveddate{XX Month, XXXX}
\reviseddate{XX Month, XXXX}
\accepteddate{XX Month, XXXX}
\publisheddate{XX Month, XXXX}
\currentdate{XX Month, XXXX}
\doiinfo{OJIM.2022.1234567}

\title{Impact of PolSAR pre-processing and balancing methods on complex-valued neural networks segmentation tasks}

\author{
    J. A. BARRACHINA \authorrefmark{1,2} \orcid{0000-0002-2139-514X},
    C. REN \authorrefmark{2} \orcid{0000-0001-8438-4539}, 
    C. MORISSEAU \authorrefmark{1},
    G. VIEILLARD \authorrefmark{1},
    AND J.-P. OVARLEZ \authorrefmark{1,2} \orcid{0000-0001-8056-4196}
}

\affil{DEMR, ONERA, Universit\'e Paris-Saclay, France}
\affil{SONDRA, CentraleSup\'elec,  Universit\'e Paris-Saclay,  91192, Gif-sur-Yvette, France}
\corresp{CORRESPONDING AUTHOR: J.-P. OVARLEZ (e-mail: jean-philippe.ovarlez@onera.fr).}
\markboth{Impact of PolSAR pre-processing and balancing methods on complex-valued neural networks segmentation tasks}{Barrachina \textit{et al.}} 


\begin{abstract}
In this paper, we investigated the semantic segmentation of \acrfull{polsar} using \acrfull{cvnn}. Although the coherency matrix is more widely used as the input of \acrshort{cvnn}, the Pauli vector has recently been shown to be a valid alternative. We exhaustively compare both methods for six model architectures, three complex-valued, and their respective real-equivalent models. We are comparing, therefore, not only the input representation impact but also the complex- against the real-valued models.
We then argue that the dataset splitting produces a high correlation between training and validation sets, saturating the task and thus achieving very high performance. We, therefore, use a different data pre-processing technique designed to reduce this effect and reproduce the results with the same configurations as before (input representation and model architectures).
After seeing that the performance per class is highly different according to class occurrences, we propose two methods for reducing this gap and performing the results for all input representations, models, and dataset pre-processing.

\end{abstract}

\begin{IEEEkeywords}
\acrshort{cvnn}, Machine Learning, \acrshort{polsar}, Radar, Semantic Segmentation
\end{IEEEkeywords}


\maketitle

\section{INTRODUCTION}
\IEEEPARstart{I}{n} the machine learning community, most neural networks are developed for processing real-valued features (voice signals, RGB images, videos, etc.). 
The signal processing community, however, is more interested in developing theories and techniques in complex fields. Indeed, complex-valued signals are encountered in various applications, such as biomedical sciences, physics, communications, and radar. 
All these fields use signal processing tools \cite{SS2010},  which are usually based on complex filtering operations and Complex-Valued representations or features (Discrete Fourier Transform, Wavelet Transform, Wiener Filtering, Matched Filter, etc.). 
Thus, \acrlong{cvnn}s (\acrshort{cvnn}s) appear as a natural choice to process and to learn from these complex-valued features since the operation performed at each layer of \acrshort{cvnn}s can be interpreted as complex filtering or multiplications. Notably, \acrshort{cvnn}s are more adapted than \acrshort{rvnn}s to extract phase information \cite{hirose2012generalization}. 
Recently, we showed that \acrshort{cvnn} are more performant in classifying non-circular Gaussian data than its real counterpart \cite{barrachina_circ_icassp}, which means \acrshort{cvnn}s are more sensible to extract phase information than \acrshort{rvnn}s. We do that by comparing vectors of random non-circular data showing that \acrshort{cvnn} can profit from this feature and extract its full potential by achieving higher accuracy, less overfitting, and lower variance than the \acrshort{rvnn}. Our findings were also cited by Reference \cite{ko2022coshnet} to justify some properties of their obtained results as they were analogous to ours.

Deep learning techniques are becoming widely popular and have extended into radar and \acrshort{polsar} image classification \cite{fix2021deep, dlr108960}.
Usually, these networks are fed with the amplitude information of the \acrshort{polsar} image while not using the phase data.

Recently, some publications started using \acrshort{cvnn}s as an alternative to conventional \acrfull{rvnn} for radar applications \cite{hirose2013complex, bassey2021survey} since radar data are generally complex-valued. 
Knowing that \acrfull{sar} data is non-circular \cite{EMGMP20, VT12} and therefore phase information play a crucial part in their representation \cite{mihai2007phase, el2013rethinking, el2015characterization}, it is no wonder that  \acrlong{cvnn}s are becoming increasingly popular for \acrshort{sar}, \acrshort{polsar} or In\acrshort{sar} applications \cite{7905999, oyama2018adaptive, gleich2018complex}.

Interestingly enough, we then propose to study the impact \acrshort{sar} pre-processing and balancing methods on the \acrshort{cvnn}s segmentation performance. Our goal is two folds first, which representation between Pauli and the coherency matrix is more relevant to train \acrshort{cvnn}s. Second, how to reduce the learning bias when the training set is imbalanced between classes. Our proposed approach is then validated on the Bretigny polarimetric \acrshort{sar} dataset. 

Section \ref{sec:related-work} summarizes the related work on the area. Sections \ref{sec:architectures} and \ref{sec:datasets} explains the model architectures and the dataset used respectively. Later, in Section \ref{sec:base-results}, we show the results for all complex models with both coherency matrix and Pauli vector input representation. Section \ref{sec:dataset-split} relaunches the simulations with a prior pre-processing, which aims to reduce the correlation between training and validation sets. Finally, Section \ref{sec:balancing} aims to reduce the gap between \acrfull{oa} and \acrfull{aa} by implementing two methods, either balancing classes through the training sampling or using a weighted loss function.



\section{RELATED WORKS} \label{sec:related-work}


Works using \acrfull{cv-cnn} have been published for \acrshort{polsar} applications. Reference \cite{zhang2017complex} compares a \acrshort{cv-cnn} with \acrshort{rv-cnn}s but lacks confidence intervals. Other recent works \cite{sun2019semi, zhao2019contrastive, zhao2019learning, superQin2021, 9034477} use a \acrshort{cv-cnn} for PolSAR applications but without comparing its result with a \acrshort{rv-cnn}.

References \cite{XIE2020255} and \cite{rs11050522} added complexity to the \acrshort{cnn} architecture by using a Recurrent \acrlong{cv-cnn} to obtain higher accuracy results.
Lately, References \cite{cao2019pixel, li2018novel} achieved \textit{state-of-the-art} performance using a \acrfull{cv-fcnn} model architecture. 
All the previously cited works of \acrshort{cvnn} applications on \acrshort{polsar} perform a pixel-wise classification task, which can be seen as a semantic segmentation task. Therefore, it is not surprising that a \acrshort{fcnn} model achieves higher accuracy as it performs semantic segmentation by design. 

\section{MODEL ARCHITECTURES} \label{sec:architectures}

\acrfull{cv-mlp} \cite{hansch2009classification}, \acrfull{cv-cnn} \cite{zhang2017complex} and \acrfull{cv-fcnn} \cite{cao2019pixel} model architectures are used for the experiments, 
These are all Complex-Valued architectures. An equivalent real-valued model architecture was also used to have the same capacity in terms of the trainable parameters as their complex-valued counterparts, as explained in References \cite{barrachina2021about, barrachina2022comparison}.
In this Section, we will give a detailed description of those models. Some slight modifications were made compared to the model's respective references with \textit{state-of-the-art} parameters not popular or known at the time of those publications. 
References \cite{hansch2009classification} and \cite{zhang2017complex} use \acrfull{sgd} as an optimizer whereas Reference \cite{cao2019pixel} use a more modern optimizer known as Adam 
which might allow models to find a lower optimal minimum. Adam was, therefore, used as the optimizer for all models. According to the results, the learning rate and momentum were tweaked for each model independently. As well as the number of epochs.

Although complex activation functions used on \acrshort{cvnn} are numerous \cite{scardapane2018complex, bassey2021survey, lee2022survey}, we will mainly focus on two types of activation functions that are an extension of the real-valued functions \cite{kuroe2003activation, barrachina2020complex}:
\begin{itemize}
	\item Type-A: $\activation_A(z) = \activation_{\Re}\left(\Re(z)\right) +       i \, \activation_{\Im}\left(\Im(z) \right)$,
	\item Type-B: $\activation_B(z) = \activation_r(|z|) \, \exp{\left( i \, \activation_\phi(\arg(z))\right)}$,
\end{itemize}
where $\activation_{\Re},\activation_{\Im},\activation_r,\activation_{\phi}$ are all real-valued functions. $\Re$ and $\Im$ operators are the real and imaginary parts of the input, respectively, and the $\arg$ operator gives the input phase.
The most popular activation functions, sigmoid, \acrfull{tanh} and \acrfull{relu}, are extensible using Type-A or Type-B approach. Although \acrshort{tanh} is already defined on the complex domain for what, its transformation is probably less interesting.
Although \cite{hansch2009classification} use $\mathrm{\tanh}$ activation function for the \acrshort{mlp} model, we decided in this work to use \acrfull{relu}. Indeed, both activation functions were tested for the \acrshort{mlp} architecture showing an interesting increase in performance when using \acrfull{relu}.
For the output layer, the \textit{softmax} activation function \cite{Goodfellow-et-al-2016} has been used.

A Normal weight initialization by K. He in \cite{he2015delving} was used, and the bias was initialized as zero. The adaptation for complex-valued weights initialization is described in Reference \cite[p.~6]{trabelsi2017deep}, which has to be done with care to keep the benefits of the K. He initialization on the complex domain.

A categorical cross-entropy loss function was used for all models. For complex models, the loss is computed twice, using first the real part and then the imaginary part as the prediction result. An average of the two error values is then calculated to be optimized. Reference \cite{cao2019pixel} defines this loss function as complex average categorical cross-entropy ($\loss^{ACE}$) which is computed as follows:
\begin{equation}
    \loss^{ACE} = \frac{1}{2} \left[ \loss^{CCE}\left( \Re{(y)}, d \right) + \loss^{CCE}\left( \Im{(y)}, d \right) \right] \, ,
\end{equation}
where $\loss^{ACE}$ is the complex average cross-entropy, $\loss^{CCE}$ is the well-known categorical cross-entropy. $y$ is the network predicted output, and $d$ is the corresponding ground truth or desired output. For real-valued output $\loss^{ACE} = \loss^{CCE}$.

Complex- and Real-Valued \acrshort{mlp} architectures had two hidden layers. For the \acrshort{cv-mlp}, 96 and 180 neurons were used for the first and second hidden layers, respectively, as presented in \cite{cao2019pixel}. The hidden layers sizes of the \acrshort{rv-mlp} were dimensioned to have the same amount of \textit{real-valued training parameters} with the same aspect ratio as explained by Reference \cite{barrachina2021about}.
The \acrshort{mlp} models presented some overfitting for what dropout with $50\%$ rate was used, which improved the performance.

Throughout literature, \acrshort{cv-cnn}s are the most popular \acrshort{cvnn} architecture used for \acrshort{polsar}. All References \cite{zhang2017complex, sun2019semi, zhao2019contrastive, zhao2019learning, superQin2021} identically dimensioned the model with the same amount of layers and kernels. Therefore, we decided to use the same architecture, which presents two convolutional layers, with 6 and 12 kernels, respectively, for the complex model. Again, their size was dimensioned for the real model as explained by Reference  \cite{barrachina2022comparison}.
All kernels were of size $3\times3$. 
Conventional arithmetic average pooling was used between both convolutional layers. The model presents a fully connected layer at the end to perform the classification.

\begin{figure}[ht]
    \centering
    \includegraphics[width=\linewidth]{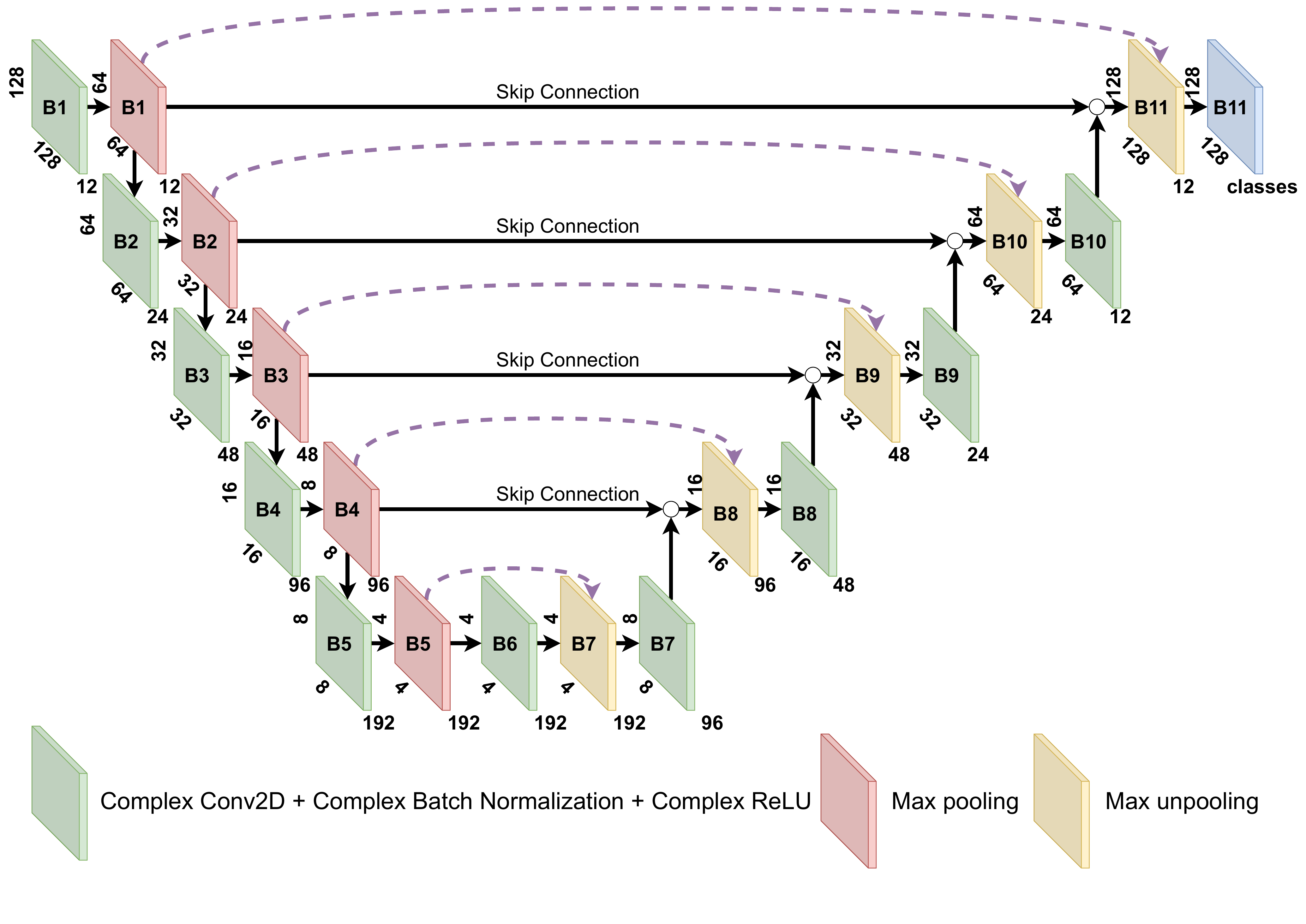}
    \caption{\acrfull{cv-fcnn} diagram.}
    \label{fig:cv-fcnn}
\end{figure}

Finally, \acrshort{cv-fcnn} (Fig. \ref{fig:cv-fcnn}) was implemented as described on \cite{cao2019pixel}, 
which is composed of the downsampling or feature extraction part and the upsampling part. The downsampling part presents several blocks (B1, B2, B3, B4, B5, and B6). Each block has two sub-modules that are represented in Fig. \ref{fig:cv-fcnn} in green and red colors. 
The upsampling part presents blocks B7, B8, B9, B10, and B11, which, in terms, are a combination of the other two sub-modules, the second one being the same green sub-module present in the downsampling section. The first sub-module (yellow) is a max-unpooling. Inspired by the functioning of Max Un-pooling explained in Reference \cite{zafar2018hands}. Max un-pooling technique receives the maxed locations of a previous Max Pooling layer and then expands an image by placing the input values on those locations and filling the rest with zeros. 


The green sub-module is a combination of a convolution layer, a \acrfull{bn} (the complex \acrshort{bn} was adapted from the real-valued \acrshort{bn} technique by Reference \cite{trabelsi2017deep}) and Complex-\acrfull{relu}. 
Reference \cite{cao2019pixel} mentions using dropout but does not indicate at which points nor their rate.
Different dropout rates were tested at several stages, such as the downsampling or upsampling part, without appreciable amelioration (and sometimes the opposite). For this reason, no Dropout was used for this model. 
This can be explained as \acrshort{bn} also acts as a regularizer, in some cases eliminating the need for Dropout \cite{srivastava2014dropout}.
The convolutional filter on each layer was of size $3\times3$, and the number used for each layer is represented in Fig. \ref{fig:cv-fcnn} for the complex model. As usual, the definition of Reference \cite{barrachina2022comparison} was used to dimension the real-valued model.

The red sub-module is a max pooling layer whose main objective is to shrink the image into smaller ones by keeping only the maximum value within a small window, in our case, of size $2\times2$. For the complex case, the absolute value of the complex number is used for comparison as proposed in \cite{zhang2017complex}. This layer complements the max-unpooling sub-module (yellow), which receives the locations where the maximum value was found. The max-unpooling layer enlarges the input image by placing their pixels according to the maxed locations received from the corresponding max-pooling layer \cite{zafar2018hands}.

The last blocks of the downsampling and upsampling parts (B6 and B11) have some differences with respect to the other blocks. B6 removes the max-pooling layer (red) completely. B11, on the other hand, replaces the \acrshort{relu} activation function with a \textit{softmax} activation function to be used for the output layer.

Each model was evaluated over around 10 Monte-Carlo trials to be able to extract statistical analysis. Simulations were done on CentraleSupélec Metz GPU servers \cite{JFix-2022}.

\section{USED DATASET} \label{sec:datasets}

\begin{figure}[ht]
         \includegraphics[width=\linewidth]{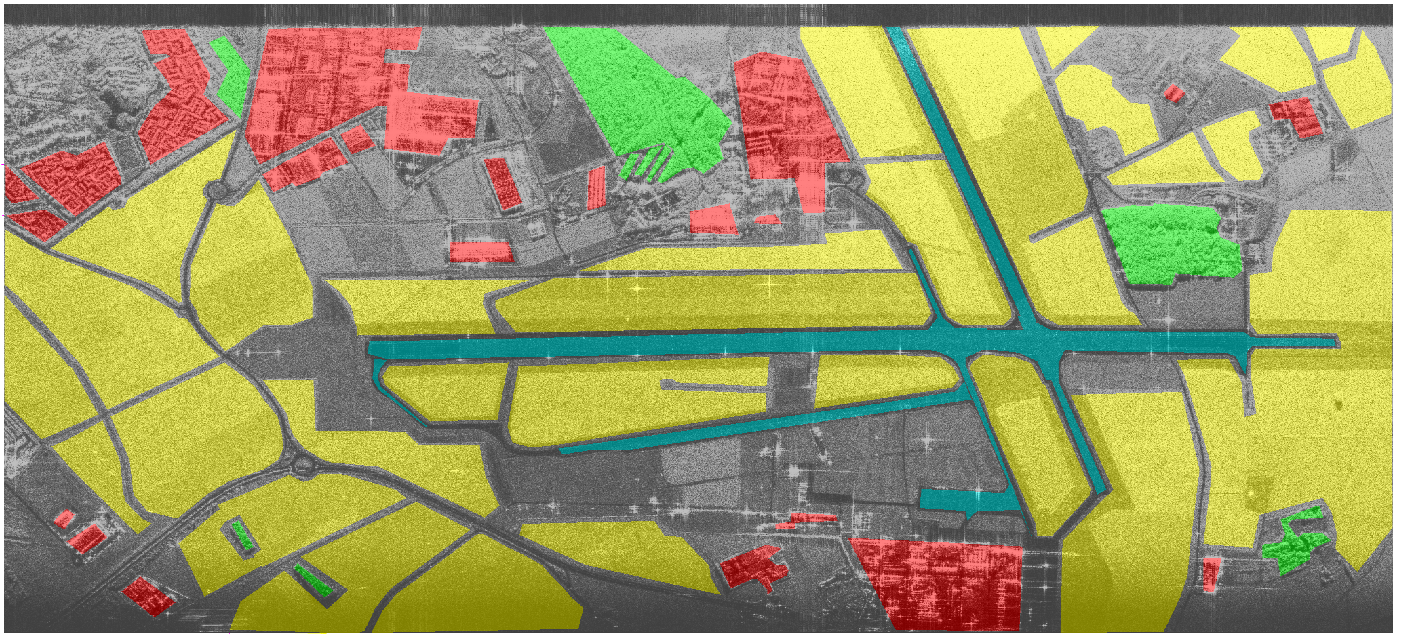}
    \caption{Bretigny image and overlapped ground truth.
    \crule{class1_br} Built-up Area; \crule{class2_br} Wood Land; \crule{class3_br} Open Area; \crule{class4_br} Runway}
    \label{fig:bret-image-and-gt}
\end{figure}

The Electromagnetic and Radar Science Department (DEMR) of ONERA, the French Aerospace Research Agency developed the RAMSES (Radar Aéroporté Multi-spectral d’Etude des Signatures) \acrshort{polsar} system in 2002 with funding from the DGA (Direction Générale de l'Armement) and CNES (Centre National d'Études Spatiales). RAMSES was developed mainly as a test bench for new technologies and to provide specific data for TDRI (Target Detection, Recognition, and Identification) algorithm evaluation. It is flown on a Transall C160 platform operated by the CEV (Centre d'Essais en Vol).

RAMSES can be configured with three bands picked among P-(430 MHz), L-(1.3 GHz), S-(3.2 GHz), C-(5.3 GHz), X-(9.5 GHz), Ku-(14.3 GHz), Ka-(35 GHz), and W-(95 GHz) bands totaling for eight different bands. From those eight, six (all but Ka and W) operate in fully polarimetric mode. The associated bandwidth and waveforms can be adjusted to meet the data acquisition objectives, and the incidence angles can be set from $30^{\circ}$ to $85^{\circ}$. The X-band and the Ku-band systems are interferometric and can collect \acrshort{polinsar} mode imagery in multi-baseline configurations, either along-track, cross-track, or both \cite{RAMSES}.


ONERA's proprietary \acrshort{polsar} image of Bretigny, France \cite{formont2010statistical} whose area is shown in Fig. \ref{fig:bret-image-and-gt}. This image was measured with RAMSES at X-band with a resolution of 1.3m. The image has a spatial resolution of 2m, an incidence angle of $30^{\circ}$, and an X frequency band. 
2,871,080 labeled images of four classes, which are Open Area ($73.20\%$), Wood Land ($5.76\%$), Built-up Area ($14.43\%$), and Runway ($6.61\%$), were manually labeled. 
However, although there was a single class for the fields (Open Area), there are different types of crops, which can impact the prediction accuracy negatively. Indeed, using k-means to split the pixels into four classes fails because it tends to group some crops with forest or runway classes \cite{mika2022}.  

\acrfull{polsar} classification algorithms generally make use of signal coherence (or equivalently phase and local phase variance) existing on a single look complex data channel vector $\mathbf{s}$ measured from two orthogonal polarimetric transmitted signals on two orthogonal polarimetric received signals. Here we use the horizontal (H) and vertical (V) polarisation, and, as with monostatic radar, the cross channels are equal; the useful received vector is: 
\begin{equation}
    \mathbf{s} = \left(S_{HH}, \sqrt{2}\,S_{HV}, S_{VV}\right)^T \, .
\end{equation}

For each pixel of the \acrfull{sar} image, this backscattering vector is usually expressed in the Pauli basis and reshaped onto one single complex vector $\in \mathbb{C}^{3}$:
\begin{equation}
    \matr{k} = \frac{1}{\sqrt{2}} \left(S_{HH}+S_{VV}, S_{HH} - S_{VV}, 2\,S_{HV}\right)^T\, .
    \label{eq:coh-matrix}
\end{equation}
The Hermitian so-called coherency matrix is then formally built according to 
\begin{equation}
    \matr{T} = \displaystyle \frac{1}{n} \sum_j^n \matr{k}_j \,\matr{k}_j^H\, ,
\end{equation}
where the operator $^H$ stands for complex conjugate transpose operation and where $n$ is the number of pixels chosen in a boxcar located in each local area of the \acrshort{sar} image. This operation is done mainly to reduce speckle noise by performing an average with the neighboring pixels.

Since $\matr{T}$ is Hermitian symmetric, its  lower triangle, excluding the diagonal, is normally discarded as it provides no additional information. As the diagonal is real-valued, the data is extended to the complex plane by adding a zero imaginary part which leads to a total of six complex values per pixel, or nine real values for the \acrshort{rvnn} architectures.

For our classification experiments, not all pixels are used for the training as the image is very large. Smaller image patches are generated using the sliding window operation \cite{rs10121984}. This method generates smaller image patches by sliding a window through the image with a given stride. The same parameters used in Reference \cite{cao2019pixel} were used for the sliding window operation method, generating images of size $12\times12$ for the \acrshort{mlp} and \acrshort{cnn} models and $128\times128$ for the \acrshort{fcnn} architecture.
References \cite{hansch2009classification} and \cite{hansch2010complex} used about 2\% of the image pixels for training whereas \cite{hou2016classification} and \cite{jiao2016wishart} used 5\%. In \cite{guo2015wishart}, the authors adopted 10\%. Finally, reference \cite{zhang2017complex} tested different sampling rates and proposed, based on the results, to use a 10\% sampling rate for both training and validation set together. For this reason, we decided to use $8\%$ and $2\%$ for training and validation, respectively, leaving the remaining pixels as the test set.


\section{COHERENCY VS. PAULI} \label{sec:base-results}

To our best knowledge, all existing work on \acrshort{polsar} \acrshort{cvnn} classification use the coherency matrix as their network input representation, except for Reference \cite{barrachina2022real}, which proposes using the Pauli vector instead under the assumption that it will work better. Indeed, the authors show that using the Pauli vector as input representation instead of the coherency matrix reduces variance and increases accuracy. However, they only perform the simulations on a \acrshort{fcnn} architecture. In this Section, we perform the same simulations on shallower \acrfull{cnn} and a \acrfull{mlp} to see if the results also stand for these models.

This increase in performance when using the Pauli vector representation is assumed to be because of the averaging operation performed on \eqref{eq:coh-matrix}, whose main objective is to reduce noise at the expense of losing resolution and mixing values of adjacent pixels, which is not favorable for pixel-wise classification. 
Although this averaging operation is done to reduce speckle noise, the averaging algorithm can be viewed as a non-trainable convolution operation on $\matr{k}\matr{k}^H$ with a constant kernel fill with $\frac{1}{n}$ values, where $n$ is the size of the kernel. Letting these kernels be trainable could enhance the performance of classification and segmentation. 

Additionally, the diagonal elements of the coherency matrix are real-valued, which is a desirable property in some instances, but that has no interest when using \acrshort{cvnn}s as they can deal with complex-valued data naturally. Therefore, they propose to use Pauli vector $\matr{k}$ as \acrshort{cvnn} input whenever this data format is available. 

\begin{table}[ht]
        \caption{Test accuracy mean results (\%)}
        \label{tab:base-results}
        \setlength{\tabcolsep}{3pt}
	\begin{tabular}{c r c c c c}
	    \toprule 
	    & & \multicolumn{2}{c}{Coherency Matrix} & \multicolumn{2}{c}{Pauli Vector} \\
	    \cmidrule(lr){3-6} \\
	    & & CV & RV & CV & RV \\
            \midrule
	    \multirow{2}{*}{CNN} &
	    \acrshort{oa} 
	    & $\textbf{95.78} \pm 0.26$
            & $94.43 \pm 0.67$
            & $95.40 \pm 0.50$
            & $94.78 \pm 0.71$ \\
	    & \acrshort{aa} 
            & $\textbf{89.72} \pm 0.67$
            & $85.82 \pm 1.53$ 
	    & $88.05 \pm 1.50$ 
	    & $86.90 \pm 1.86$ \\
	    \midrule
	    \multirow{2}{*}{MLP} &
	    \acrshort{oa} 
	    & $95.09 \pm 0.02$ 
	    & $\textbf{95.13} \pm 0.01$
            & $88.55 \pm 0.04$
            & $87.77 \pm 0.04$  \\
	    & \acrshort{aa} 
            & $87.10 \pm 0.15$
            & $\textbf{88.40} \pm 0.09$
	    & $64.69 \pm 0.08$ 
	    & $63.13 \pm 0.10$ \\
        \bottomrule
	\end{tabular}
\end{table}

The results for the \acrshort{fcnn} architecture were already published on \cite{barrachina2022complex}, where \acrshort{cv-fcnn} achieved a high of $99.83 \pm 0.02 \%$ \acrshort{oa} and $98.69 \pm 0.33 \%$ \acrshort{aa} and \acrshort{rv-fcnn} $99.69 \pm 0.06 \%$ and $98.62 \pm 0.20 \%$ respectively, all using Pauli vector as the input representation and even lower for the coherency matrix representation. 
We, therefore, performed the same simulations for the other two architectures, whose results are shown in Table \ref{tab:base-results}. We can verify that although \acrshort{fcnn} obtain better results when using the Pauli vector as input representation, this is not the case with \acrshort{cnn} and \acrshort{mlp} models. 
This can be explained by the fact that \acrshort{mlp} have no easy way to deal with the speckle noise, as their operation is not analogous to that of a filter. For this model, it is logical to assume that the speckle noise reduction out-weights the loss of information caused by the average filter. 
For \acrshort{cnn}, the complex model obtained higher performance with the coherency matrix, although further simulations should be done as confidence intervals intersect. \acrshort{rv-mlp}, on the other hand, favored the Pauli vector.
Indeed, by using a complex model, the lower amount of filters, with the fact that we are using only two convolutional layers, may not always suffice to reduce speckle noise while extracting the pertinent features, penalizing performance. Indeed, under these characteristics, \acrshort{cv-cnn} have a lower degree of freedom than \acrshort{rv-mlp} \cite{hirose2011nature}, although generally helps achieve better results, in this case, it prevents the model from generalizing better. A deeper model with more filters per layer should be used to verify this hypothesis.

Another surprise when using \acrshort{mlp} architecture is that, unlike for \acrshort{fcnn} and \acrshort{cnn}, the real-valued model outperformed the complex one, although by a very small margin as, for example, \acrshort{cv-mlp} upper mean \acrshort{oa} estimate ($95.11\%$) almost coincide with \acrshort{rv-mlp} lower mean estimate ($95.12\%$). This fact remains to be explored. 

In general, although using a complex-valued model might be significant to increase performance, the used model architecture has a bigger relevance and impact in the result, with \acrshort{fcnn} outperforming \acrshort{cnn} and \acrshort{cnn} outperforming \acrshort{mlp} regardless of the architecture data type (complex or real).


\section{DATASET SPLIT} \label{sec:dataset-split}

In the previous Section, results got as high as $99.83\%$ mean test \acrlong{oa} for \acrshort{fcnn} architecture, indicating that the problem is over-saturated.
Reference \cite{barrachina2022complex} states that this issue is mainly due to a close correlation between the training, validation, and test sets. Therefore, they propose to split the dataset to reduce this effect. Indeed, with this pre-processing, the \acrshort{oa} accuracy drops significantly to under $94\%$. We, therefore, perform the same dataset splitting and relaunch the simulations for all models to verify the impact of this training and validation correlation.

\begin{figure}[ht]
\centering
\subfloat[train\label{fig:bret-train}]{
  \includegraphics[width=.65\linewidth]{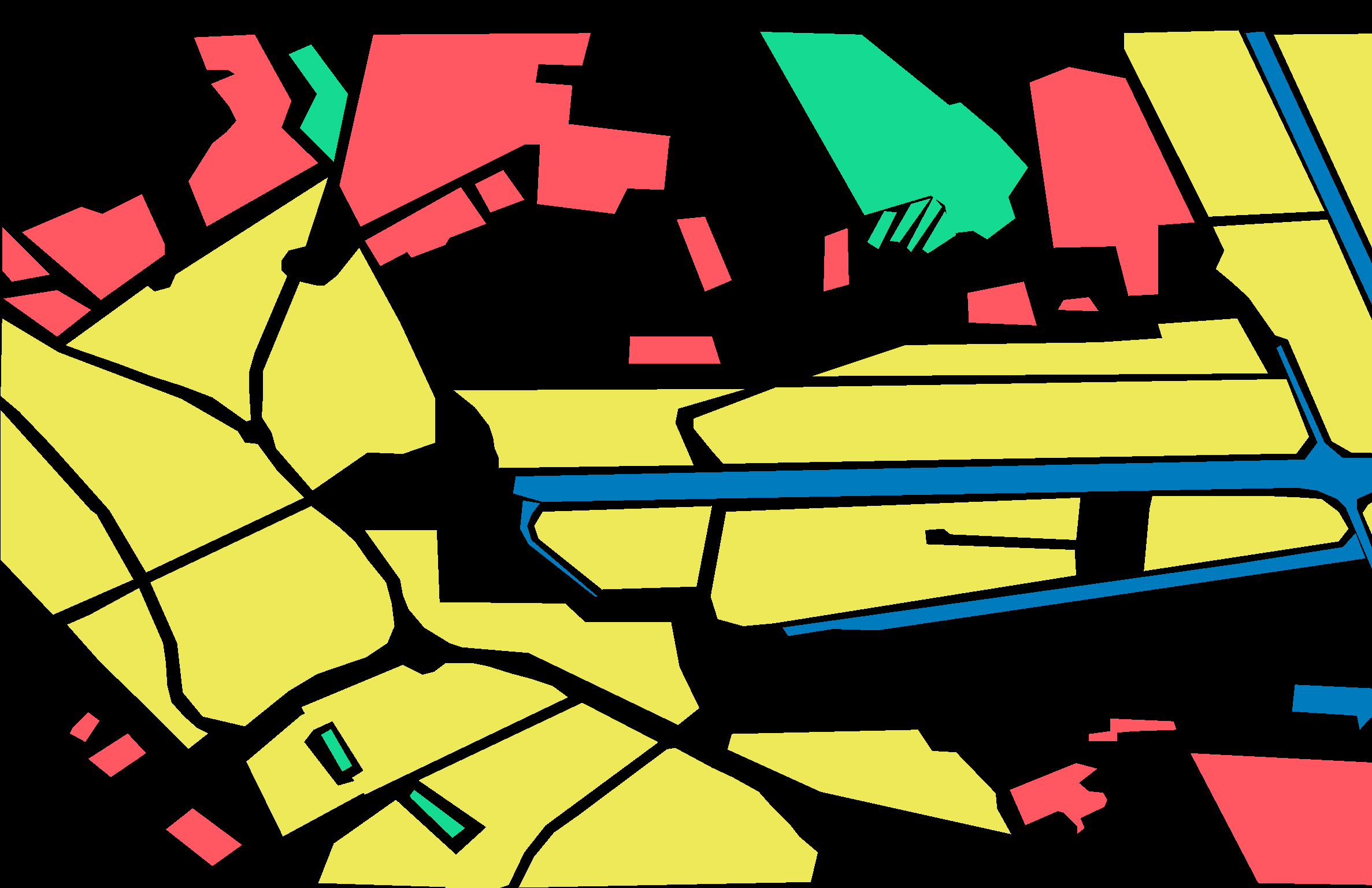}}
\subfloat[val\label{fig:bret-val}]{
  \includegraphics[width=.14\linewidth]{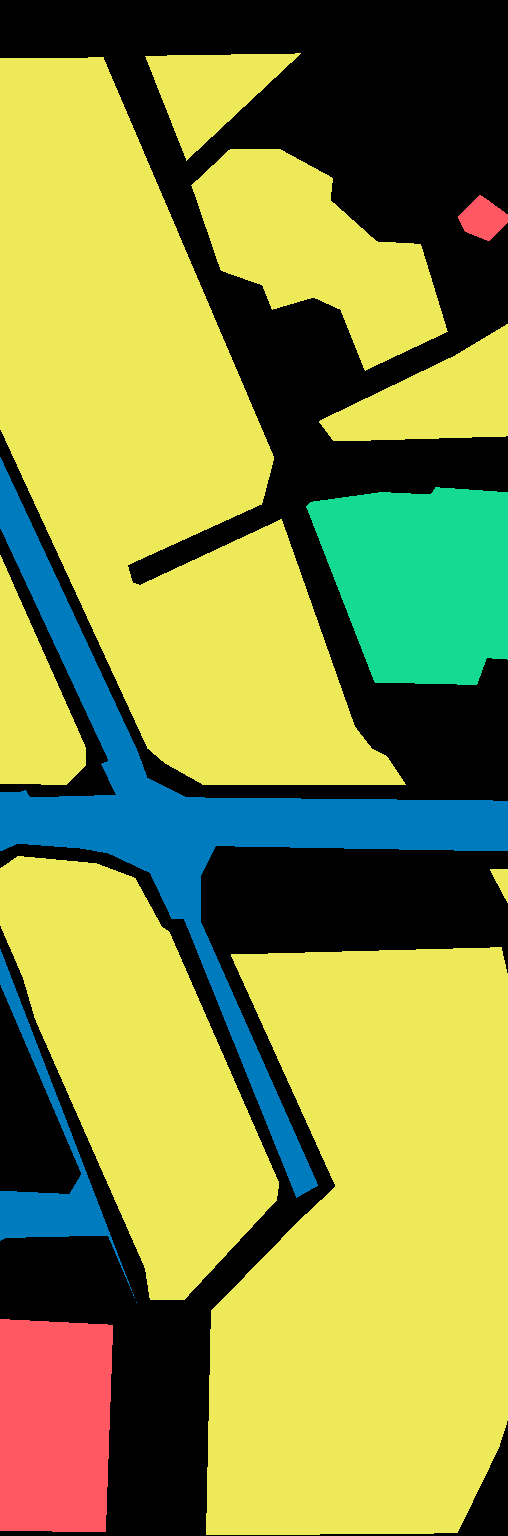}}
\subfloat[test\label{fig:bret-test}]{
  \includegraphics[width=.14\linewidth]{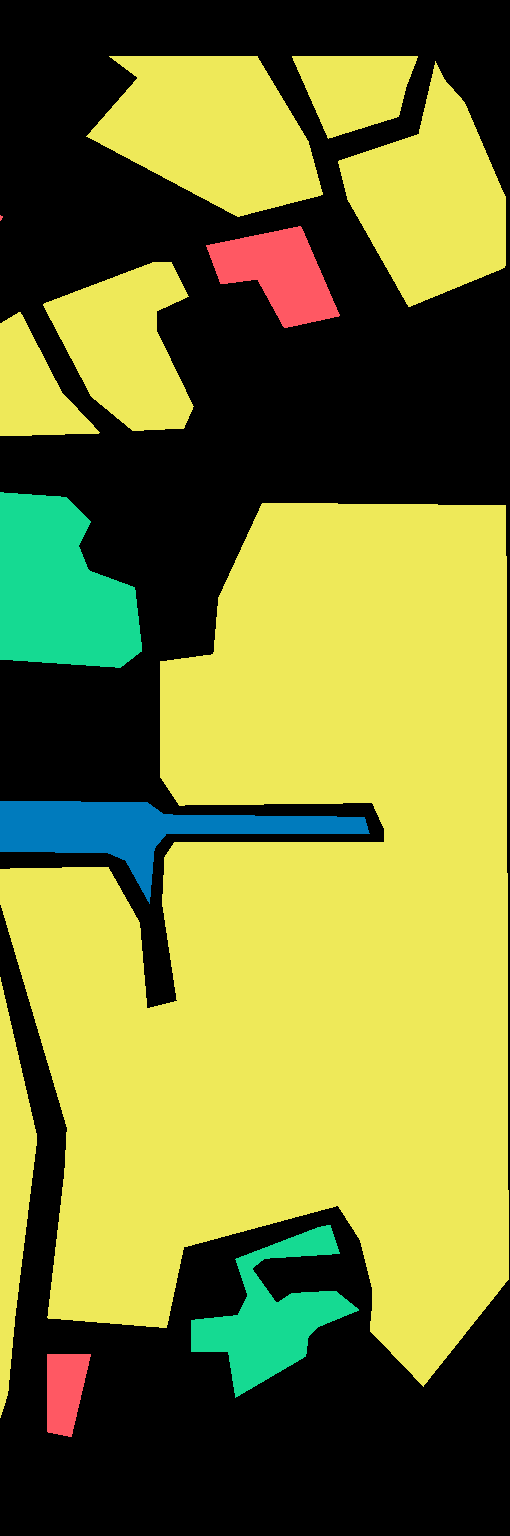}}
\caption{Split of Bretigny dataset; $70\%$ as the training set, $15\%$ as the validation set, and $15\%$ as the test set. \crule{class1_br} Built-up Area; \crule{class2_br} Wood Land; \crule{class3_br} Open Area; \crule{class4_br} Runway}
\label{fig:bret-split-dataset}
\end{figure}

The dataset was split as shown in Fig. \ref{fig:bret-split-dataset}. $70\%$ of the image was used as a training set, and $15\%$ was used for both validation and test set. Note that the four classes are present in each sub-image as shown in Fig. \ref{fig:bret-split-dataset}. 
This method not only avoids the ground-truth overlap but also prevents pixels from the same class from being close to each other.

\begin{table}[ht]
	\centering
        \caption{Test accuracy mean results with dataset split method (\%)}
        \label{tab:dataset-split}
        \setlength{\tabcolsep}{3pt}
	\begin{tabular}{c r c c c c}
	    \toprule 
	    & & \multicolumn{2}{c}{Coherency Matrix} & \multicolumn{2}{c}{Pauli Vector} \\
	    \cmidrule(lr){3-6} \\
	    & & CV & RV & CV & RV \\
            \midrule
	    \multirow{2}{*}{CNN} &
	    \acrshort{oa} 
	    & $\textbf{93.92} \pm 0.10$
            & $93.31 \pm 0.10$
            & $93.03 \pm 0.18$ 
            & $92.52 \pm 0.24$ \\
	    & \acrshort{aa} 
            & $\textbf{90.76} \pm 0.16$
            & $90.19 \pm 0.14$ 
	    & $87.76 \pm 0.37$ 
	    & $86.80 \pm 0.40$ \\
	    \midrule
	    \multirow{2}{*}{MLP} &
	    \acrshort{oa} 
	    & $\textbf{93.40} \pm 0.06$ 
	    & $92.34 \pm 0.16$
            & $81.47 \pm 0.47$
            & $82.29 \pm 0.35$  \\
	    & \acrshort{aa} 
            & $\textbf{89.16} \pm 0.10$
            & $\textbf{89.16} \pm 0.13$
	    & $74.29 \pm 0.43$ 
	    & $74.75 \pm 0.31$ \\
        \bottomrule
	\end{tabular}
\end{table}

Again, the results for the \acrshort{fcnn} architecture were published on \cite{barrachina2022complex}, where \acrshort{cv-fcnn} achieved a high of $93.62 \pm 0.20 \%$ \acrshort{oa} and $75.31 \pm 0.63 \%$ \acrshort{aa} and \acrshort{rv-fcnn} $92.63 \pm 0.29 \%$ and $76.20 \pm 0.80 \%$ respectively. Resulting in a significant decrease from previous results. These results meant that this technique successfully unsaturated the segmentation task to a harder case.
Table \ref{tab:dataset-split} shows the test accuracy results when using the splitting method. We can see that this split has a smaller impact on \acrshort{cnn} and \acrshort{mlp} models compared with \acrshort{fcnn}. 
This is probably due to the fact that this method reduces the total image patches for training for the \acrshort{fcnn} models while not affecting the same figure for the \acrshort{cnn} or \acrshort{mlp} models.



\section{BALANCING DATASETS} \label{sec:balancing}


Previous results show a big gap between \acrlong{oa} and \acrlong{aa} from around $3\%$ when using the coherency matrix and even more than $6\%$ when using the Pauli vector. This dissimilarity can be explained by a huge difference in class occurrences.

We decided to reduce the gap between both metrics by either balancing the train and validation set using different sampling per class or using a weighted loss that heavily penalizes the classes with fewer occurrences. For the last case, the loss per label was multiplied $n_l/n_c$ where $n_l$ is the number of samples of less frequent class, and $n_c$ is the total occurrences of the corresponding label.

Balancing classes using the sampling method is easy for \acrshort{mlp} and \acrshort{cnn} models as it suffices to take the same amount of central pixels for each class. Also, as a low percentage of pixels was used for train and validation (of $10\%$), this method did not affect the total number of image patches used for training and validation sets.

However, this technique is not suitable for the \acrshort{fcnn} model, the results did not work as expected, having little impact on the network's performance. 
Notably, when using the sliding window operation for the dataset split method with a stride of $25$ and images of size $128\times128$, some generated small sub-images have only one class present in them (shown in Table \ref{tab:train-occ} in the column \textit{Single-class images}). 
Randomly removing pixels to have the same amount of pixel classes will have little impact on the total amount of single images, so class Open Area will still have a relation of almost 10:1 compared with Forest images and even 20:1 with Runway images. The only effect it will have is that those Open Area single-class images will have fewer pixels, losing generality. Under this condition, with a batch size of 32, Runway images have an average of 1 image per batch, meaning that batches, where this class is not present will indeed occur.

\begin{table}[h]
    \centering
    \caption{Training set class occurrences with dataset split method}
    \setlength{\tabcolsep}{3pt}
    \label{tab:train-occ}
    \begin{tabular}{r c c c c}
         \toprule
         \multirow{ 2}{*}{Class} & Total  & Pixels in  & Single-class & Mixed  \\ 
         & Pixels & mixed images & images & Images \\
         \midrule
         Forest & 2,721,238 & 592,309 & 264 & 304 \\
         Runway & 3,091,975 & 2,900,017 & 109 & 931 \\
         Built-Up Area & 9,750,257 & 2,620,141 & 1134 & 686 \\
         Open Area & 33,245,055 & 10,797,471 & 2045 & 1519 \\
         \bottomrule
    \end{tabular}
\end{table}

To balance this dataset, the following algorithm was implemented.
As a first step, single-class images are removed to get the same number of images that contains each class (counting both single-class images and mixed images). In our case, this is not possible to achieve because we are in the less likely case where class 0 (Forest) has a total of 568 images, and even by deleting all single-class images of other classes, they will still have more presence in images. Therefore, in this step, all single-class images for every class except Forest were deleted. Under this scenario, the label occurrences will now be as shown in figure \ref{tab:train-occ-2}. Note that Forest is no longer the less frequent class, but now this role was taken by the Build-up Area. Also, only with this step all classes except the Open Area have a very similar occurrence.
In cases where not all single-class images need to be removed (not present in our example), the images will not be removed randomly, but the images with fewer pixels will be removed.

\begin{table}[h]
    \centering
    \caption{Training set class occurrences with dataset split method after balancing single-class images.}
    \setlength{\tabcolsep}{3pt}
    \label{tab:train-occ-2}
    \begin{tabular}{r c c c c}
         \toprule
         \multirow{ 2}{*}{Class} & Total  & Pixels in  & Single-class & Mixed  \\ 
         & Pixels & mixed images & images & Images \\ 
         \midrule
         Forest & 2,721,238 & 592,309 & 264 & 304 \\
         Runway & 2,900,017 & 2,900,017 & 0 & 931 \\
         Urban & 2,620,141 & 2,620,141 & 0 & 686 \\
         Open Area & 10,797,471 & 10,797,471 & 0 & 1519 \\
         \bottomrule
    \end{tabular}
\end{table}

The second step will be to remove pixels to balance the total number of pixels; the final code balancing will therefore consist of two phases, image patches balancing followed by pixel-wise balancing as shown in the following pseudo-code 

\begin{lstlisting}[language=python]
def balance_patches(patches, labels):
    patches, labels = _remove_exceeding_one_class_images(patches, labels)
    # Only labels can be used here, as the input should not be changed.
    labels = _balance_total_pixels_of_patch(labels)
    return patches, labels
\end{lstlisting}

However, instead of randomly removing the exceeding pixels, it would be better to remove pixels from images that have more class occurrences. For example, if ten labels have to be removed and there are only two images containing that class, one with ten occurrences and the other with $100$ occurrences, it will be preferable to remove all ten pixels from the image that has $100$ occurrences instead of deleting five from each.

In this step, by knowing the total pixels we need to get (in our example, $2,620,141$) and knowing each class total number of images, we can know the average of pixels these images should have. If all images have more than this average, the balancing will be simply to remove pixels so that each image has this average. 

If, on the contrary, some images have fewer pixels than the expected average, there will forcibly be others that have more. In this case, per class, images are ordered in ascending order. 
The first images will have a lower total number of pixels than the expected average. For this reason, they will not be changed, and the total amount of pixels to achieve (in our example, $2,620,141$) will be updated by subtracting the pixels present in the current image. These images will not be counted when computing the new average, for the expected average will increase until there is a moment when the total pixels of the image will meet or be higher than the expected average. In this case, labeled pixels of the current class will be randomly removed so that it has the same pixels as the intended average, this will be repeated for all the following images. 
In the example of having only two images, a total of $100$ pixels are needed, but both images sum $110$ pixels. The average number of pixels per image will therefore be $55$, which is higher than the total of pixels of the first image (10). The total number of pixels to be achieved will therefore be $90$ (100 - 10). In this case, with only one image remaining, the total average will also be 90, meaning that 10 pixels should be removed from the last image.

Table \ref{tab:train-occ-3} shows the final numbers of pixels after the balancing algorithm is implemented. 
In the following, there is the pseudo-code used for this balancing part. 
The final code used for balancing the classes can be found in \href{https://github.com/NEGU93/CVNN-PolSAR/blob/ee51db13bca1b4c11b2e03592d5534f95d076f41/src/dataset_reader.py#L942}{\path{github.com/NEGU93/CVNN-PolSAR}}.

\begin{lstlisting}[language=python]
def _balance_total_pixels_of_patch(label_patches):
    to_be_achieved = np.min(total_pixels_per_class_list)
    for cls in range(label_patches.shape[-1]):
        tmp_to_be_achieved = to_be_achieved
        # Get the indexes of all images that contain class cls in ascending order
        ordered_images = get_ordered_list_indexes(label_patches, cls)
        # Get total images that contain class cls
        total_images = get_total_images(label_patches, cls)
        for image_index in ordered_images:
            occ = occurrences(label_patches)
            avg = tmp_to_be_achieved / total_images
            total_images -= 1
            if occ <= avg:
                to_be_achieved -= occ
            else:
                label_patches[image_index] = _randomly_remove(label_patches[image_index], number_of_pixels_to_remove=occ - avg)
                to_be_achieved -= avg
    return label_patches
\end{lstlisting}

\begin{table}[h]
    \centering
    \caption{Training set class occurrences with dataset split method after dataset balancing.}
    \setlength{\tabcolsep}{3pt}
    \label{tab:train-occ-3}
    \begin{tabular}{r c c c c}
         \toprule
         \multirow{ 2}{*}{Class} & Total  & Pixels in  & Single-class & Mixed  \\ 
         & Pixels & mixed images & images & Images \\
         \midrule
         Forest & 2,620,141 & 592,309 & 264 & 304 \\
         Runway & 2,620,141 & 2,620,141 & 0 & 931 \\
         Urban & 2,620,141 & 2,620,141 & 0 & 686 \\
         Open Area & 2,620,141 & 2,620,141 & 0 & 1519 \\
         \bottomrule
    \end{tabular}
\end{table}

Results for both techniques, dataset balancing and weighted loss, where tested with both the standard dataset pre-processing of Reference \cite{cao2019pixel}, used in Section \ref{sec:base-results} and the dataset splitting proposed by Reference \cite{barrachina2022complex} and explained in Section \ref{sec:dataset-split}.
\acrshort{fcnn} did not obtain good results for weighted loss balancing for what results were omitted, the reason why it happened should be revised, but it may be for the same reason that randomly removing pixels from the image did not work.

\begin{figure}
    \centering
    \begin{tikzpicture}

\definecolor{color0}{rgb}{0.086,0.858,0.576}
\definecolor{color1}{rgb}{0,0.486,0.745}
\definecolor{color2}{rgb}{1,0.349,0.392}
\definecolor{color3}{rgb}{0.937,0.917,0.352}

\begin{axis}[
tick align=outside,
tick pos=left,
x grid style={white!69.0196078431373!black},
xmin=-0.5525, xmax=5.5525,
xtick style={color=black},
xtick={0.2,1.2,2.2,3.2,4.2,5.2},
xticklabels={
  C,
  C-DB,
  C-WL,
  R,
  R-DB,
  R-WL
},
width=\linewidth,
height=0.5\linewidth,
y grid style={white!69.0196078431373!black},
ylabel={Train Accuracy \%},
ymajorgrids,
ymin=0.7, ymax=1,
yminorgrids,
ytick style={color=black},
ytick={0,0.2,0.4,0.6, 0.7,0.8, 0.9,1},
yticklabels={0,0.2,0.4,0.6, 0.7,0.8, 0.9,1}
]
\draw[draw=none,fill=color0] (axis cs:-0.275,0) rectangle (axis cs:-0.125,0.795719021765243);
\draw[draw=none,fill=color0] (axis cs:0.725,0) rectangle (axis cs:0.875,0.955304716481187);
\draw[draw=none,fill=color0] (axis cs:1.725,0) rectangle (axis cs:1.875,0.948551386152885);
\draw[draw=none,fill=color0] (axis cs:2.725,0) rectangle (axis cs:2.875,0.77093884665666);
\draw[draw=none,fill=color0] (axis cs:3.725,0) rectangle (axis cs:3.875,0.942235390470685);
\draw[draw=none,fill=color0] (axis cs:4.725,0) rectangle (axis cs:4.875,0.928499491599283);
\draw[draw=none,fill=color1] (axis cs:-0.141666666666667,0) rectangle (axis cs:0.00833333333333333,0.987222494286647);
\draw[draw=none,fill=color1] (axis cs:0.858333333333333,0) rectangle (axis cs:1.00833333333333,0.996618008382714);
\draw[draw=none,fill=color1] (axis cs:1.85833333333333,0) rectangle (axis cs:2.00833333333333,0.996506692784851);
\draw[draw=none,fill=color1] (axis cs:2.85833333333333,0) rectangle (axis cs:3.00833333333333,0.983620618895182);
\draw[draw=none,fill=color1] (axis cs:3.85833333333333,0) rectangle (axis cs:4.00833333333333,0.995215108156285);
\draw[draw=none,fill=color1] (axis cs:4.85833333333333,0) rectangle (axis cs:5.00833333333333,0.995054584256751);
\draw[draw=none,fill=color2] (axis cs:-0.00833333333333336,0) rectangle (axis cs:0.141666666666667,0.920893385131462);
\draw[draw=none,fill=color2] (axis cs:0.991666666666667,0) rectangle (axis cs:1.14166666666667,0.919704196174784);
\draw[draw=none,fill=color2] (axis cs:1.99166666666667,0) rectangle (axis cs:2.14166666666667,0.919639092560428);
\draw[draw=none,fill=color2] (axis cs:2.99166666666667,0) rectangle (axis cs:3.14166666666667,0.917185851445453);
\draw[draw=none,fill=color2] (axis cs:3.99166666666667,0) rectangle (axis cs:4.14166666666667,0.915444428385605);
\draw[draw=none,fill=color2] (axis cs:4.99166666666667,0) rectangle (axis cs:5.14166666666667,0.919563726524583);
\draw[draw=none,fill=color3] (axis cs:0.125,0) rectangle (axis cs:0.275,0.990427928364552);
\draw[draw=none,fill=color3] (axis cs:1.125,0) rectangle (axis cs:1.275,0.957333911451559);
\draw[draw=none,fill=color3] (axis cs:2.125,0) rectangle (axis cs:2.275,0.952620416645665);
\draw[draw=none,fill=color3] (axis cs:3.125,0) rectangle (axis cs:3.275,0.988350038012195);
\draw[draw=none,fill=color3] (axis cs:4.125,0) rectangle (axis cs:4.275,0.952142409789469);
\draw[draw=none,fill=color3] (axis cs:5.125,0) rectangle (axis cs:5.275,0.949229283353762);
\draw (axis cs:-0.25,0.795719021765243) node[
  scale=0.5,
  anchor=north east,
  text=black,
  rotate=90.0
]{\bfseries 79.57\%};
\draw (axis cs:0.75,0.955304716481187) node[
  scale=0.5,
  anchor=north east,
  text=black,
  rotate=90.0
]{\bfseries 95.53\%};
\draw (axis cs:1.75,0.948551386152885) node[
  scale=0.5,
  anchor=north east,
  text=black,
  rotate=90.0
]{\bfseries 94.86\%};
\draw (axis cs:2.75,0.77093884665666) node[
  scale=0.5,
  anchor=north east,
  text=black,
  rotate=90.0
]{\bfseries 77.09\%};
\draw (axis cs:3.75,0.942235390470685) node[
  scale=0.5,
  anchor=north east,
  text=black,
  rotate=90.0
]{\bfseries 94.22\%};
\draw (axis cs:4.75,0.928499491599283) node[
  scale=0.5,
  anchor=north east,
  text=black,
  rotate=90.0
]{\bfseries 92.85\%};
\draw (axis cs:-0.116666666666667,0.987222494286647) node[
  scale=0.5,
  anchor=north east,
  text=black,
  rotate=90.0
]{\bfseries 98.72\%};
\draw (axis cs:0.883333333333333,0.996618008382714) node[
  scale=0.5,
  anchor=north east,
  text=black,
  rotate=90.0
]{\bfseries 99.66\%};
\draw (axis cs:1.88333333333333,0.996506692784851) node[
  scale=0.5,
  anchor=north east,
  text=black,
  rotate=90.0
]{\bfseries 99.65\%};
\draw (axis cs:2.88333333333333,0.983620618895182) node[
  scale=0.5,
  anchor=north east,
  text=black,
  rotate=90.0
]{\bfseries 98.36\%};
\draw (axis cs:3.88333333333333,0.995215108156285) node[
  scale=0.5,
  anchor=north east,
  text=black,
  rotate=90.0
]{\bfseries 99.52\%};
\draw (axis cs:4.88333333333333,0.995054584256751) node[
  scale=0.5,
  anchor=north east,
  text=black,
  rotate=90.0
]{\bfseries 99.51\%};
\draw (axis cs:0.0166666666666666,0.920893385131462) node[
  scale=0.5,
  anchor=north east,
  text=black,
  rotate=90.0
]{\bfseries 92.09\%};
\draw (axis cs:1.01666666666667,0.919704196174784) node[
  scale=0.5,
  anchor=north east,
  text=black,
  rotate=90.0
]{\bfseries 91.97\%};
\draw (axis cs:2.01666666666667,0.919639092560428) node[
  scale=0.5,
  anchor=north east,
  text=black,
  rotate=90.0
]{\bfseries 91.96\%};
\draw (axis cs:3.01666666666667,0.917185851445453) node[
  scale=0.5,
  anchor=north east,
  text=black,
  rotate=90.0
]{\bfseries 91.72\%};
\draw (axis cs:4.01666666666667,0.915444428385605) node[
  scale=0.5,
  anchor=north east,
  text=black,
  rotate=90.0
]{\bfseries 91.54\%};
\draw (axis cs:5.01666666666667,0.919563726524583) node[
  scale=0.5,
  anchor=north east,
  text=black,
  rotate=90.0
]{\bfseries 91.96\%};
\draw (axis cs:0.15,0.990427928364552) node[
  scale=0.5,
  anchor=north east,
  text=black,
  rotate=90.0
]{\bfseries 99.04\%};
\draw (axis cs:1.15,0.957333911451559) node[
  scale=0.5,
  anchor=north east,
  text=black,
  rotate=90.0
]{\bfseries 95.73\%};
\draw (axis cs:2.15,0.952620416645665) node[
  scale=0.5,
  anchor=north east,
  text=black,
  rotate=90.0
]{\bfseries 95.26\%};
\draw (axis cs:3.15,0.988350038012195) node[
  scale=0.5,
  anchor=north east,
  text=black,
  rotate=90.0
]{\bfseries 98.84\%};
\draw (axis cs:4.15,0.952142409789469) node[
  scale=0.5,
  anchor=north east,
  text=black,
  rotate=90.0
]{\bfseries 95.21\%};
\draw (axis cs:5.15,0.949229283353762) node[
  scale=0.5,
  anchor=north east,
  text=black,
  rotate=90.0
]{\bfseries 94.92\%};
\end{axis}

\end{tikzpicture}
    \caption[CNN Coherency matrix results per class]{CNN Coherency matrix results per class. Model C: \acrshort{cv-cnn}; C-DB: \acrshort{cv-cnn} dataset balanced; C-WL: \acrshort{cv-cnn}; R: \acrshort{rv-cnn}; weighted loss; R-DB: \acrshort{rv-cnn} dataset balanced; R-WL: \acrshort{rv-cnn} weighted loss}
    \label{fig:cnn-bars}
\end{figure}
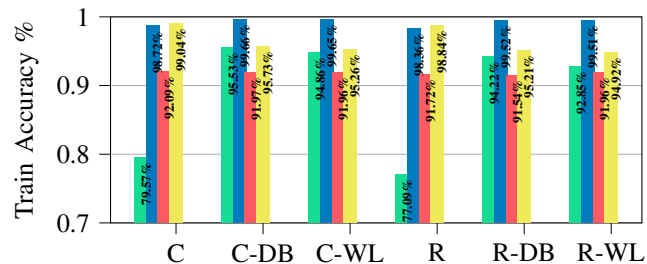

Fig. \ref{fig:cnn-bars} shows the mean test accuracy per class for all \acrshort{cnn} models with coherency matrix as input (a similar graph is obtained when using the Pauli vector representation). It can be seen how both the dataset balance and the weighted loss method obtain a higher accuracy for the Forest class while reducing the Open Field accuracy, thus, obtaining higher \acrshort{aa} at the cost of reducing the \acrshort{oa}. It is also important to note that the complex-valued model acquired higher accuracy than their real-valued equivalent model for every class category.

Results can be seen in Tables \ref{tab:balanced-random} and \ref{tab:balanced-split}.
In general, as also shown in Fig. \ref{fig:cnn-bars}, both techniques successfully reduced the gap between \acrshort{oa} and \acrshort{aa} results. However, in most cases, dataset balancing worked better by obtaining a smaller gap and higher accuracy values. 

For \acrshort{fcnn}, the dataset balance without the dataset splitting had a negative effect on the accuracy, with both \acrshort{oa} and \acrshort{aa} having less accuracy than before. The balancing technique reduced too much the size of the training data meaning that, although the accuracy per class was more stable, the accuracy dropped in general. The clearest sign of this case was the \acrshort{rv-fcnn} with the coherency matrix, which had an accuracy of around $50\%$. 
For \acrshort{cnn}, the balancing method worked very well with \acrshort{aa} some times even higher than the \acrshort{oa}. This time, Pauli input representation achieved higher results than the coherency matrix.
It is important to note that for the coherency matrix with no dataset splitting, \acrshort{cv-cnn} and \acrshort{rv-cnn} obtained almost identical \acrshort{oa}. However, this result was not statistically significant in favor of \acrshort{rv-cnn} added to the fact that \acrshort{cv-cnn} did obtained a higher \acrshort{aa}.

As before, \acrshort{fcnn} obtained higher results when using the Pauli vector representation as input, whereas \acrshort{mlp} obtained higher results with the Coherency matrix. For \acrshort{cnn}, however, it was not clear which input representation was best with a small tendency towards coherency matrix.

\begin{table}[ht]
	\centering
        \caption{Test accuracy mean results for both dataset balancing and weighted loss (\%)}
        \label{tab:balanced-random}
        \setlength{\tabcolsep}{3pt}
	\begin{tabular}{c r c c c c}
	    \toprule 
            \multicolumn{2}{c}{Dataset} & \multicolumn{2}{c}{Coherency Matrix} 
            & \multicolumn{2}{c}{Pauli Vector} \\
	    \cmidrule(lr){3-6} \\
	    & & CV & RV & CV & RV \\
	    \midrule
	    \multirow{2}{*}{FCNN} &
	    \acrshort{oa} 
	    & $83.08 \pm 1.80$ 
            & $46.14 \pm 3.41$ 
            & $\textbf{98.85} \pm 0.07$
	    & $98.50 \pm 0.13$  \\
	    & \acrshort{aa} 
	    & $69.45 \pm 2.90$ 
            & $55.73 \pm 2.82$ 
	    & $\textbf{98.17} \pm 0.12$
            & $98.04 \pm 0.27$ \\
	    \midrule
	    \multirow{2}{*}{CNN} &
	    \acrshort{oa} 
	    & $94.41 \pm 0.06$  
            & $94.42 \pm 0.09$ 
	    & $\textbf{94.83} \pm 0.11$ 
            & $94.60 \pm 0.10$  \\
	    & \acrshort{aa}  
            & $94.84 \pm 0.06$  
	    & $94.36 \pm 0.06$ 
            & $\textbf{95.42} \pm 0.06$ 
            & $95.25 \pm 0.04$  \\
	    \midrule
	    \multirow{2}{*}{MLP} &
	    \acrshort{oa}
	    & $92.77 \pm 0.11$  
            & $\textbf{92.82} \pm 0.16$ 
	    & $71.70 \pm 0.09$  
            & $71.84 \pm 0.10$   \\
	    & \acrshort{aa}       
            & $92.38 \pm 0.03$  
	    & $\textbf{92.85} \pm 0.04$ 
            & $81.13 \pm 0.06$
            & $80.56 \pm 0.10$  \\
            \hline \hline 
            \\\\[-1.\medskipamount]
            \multicolumn{2}{c}{Loss} & \multicolumn{2}{c}{Coherency Matrix} 
            & \multicolumn{2}{c}{Pauli Vector} \\
	    \cmidrule(lr){3-6} \\
	    & & CV & RV & CV & RV \\
	    \midrule
	    \multirow{2}{*}{CNN} &
	    \acrshort{oa} 
	    & $90.66 \pm 0.48$ 
            & $87.11 \pm 1.28$
	    & $\textbf{92.35} \pm 0.81$ 
            & $91.61 \pm 1.11$ \\
	    & \acrshort{aa} 
	    & $89.96 \pm 0.30$ 
            & $86.31 \pm 0.82$ 
	    & $\textbf{92.12} \pm 1.06$
            & $91.53 \pm 1.17$ \\
	    \midrule
	    \multirow{2}{*}{MLP} &
	    \acrshort{oa} 
	    & $\textbf{93.40} \pm 0.13$ 
            & $92.70 \pm 0.20$
	    & $71.96 \pm 0.32$
            & $71.51 \pm 0.22$ \\
	    & \acrshort{aa} 
	    & $91.05 \pm 0.08$ 
            & $\textbf{91.39} \pm 0.08$ 
	    & $79.27 \pm 0.06$ 
            & $78.04 \pm 0.01$ \\
        \bottomrule
	\end{tabular}
\end{table}

For the dataset balance with splitting, \acrshort{cv-fcnn} obtained the higher \acrshort{oa} but, although the dataset balancing did increased the \acrshort{aa} by around $5\%$, it was \acrshort{cv-cnn} which obtained the higher \acrshort{aa}. 
This time, \acrshort{cv-cnn} obtained higher \acrshort{oa} when using the Pauli vector as input representation but higher \acrshort{aa} when using the coherency matrix.

\begin{table}[ht]
	\centering
        \caption{Test accuracy mean results for both dataset balancing and weighted loss with split method (\%)}
        \label{tab:balanced-split}
        \setlength{\tabcolsep}{3pt}
	\begin{tabular}{c r c c c c}
	    \toprule 
            \multicolumn{2}{c}{Dataset} & \multicolumn{2}{c}{Coherency Matrix} & \multicolumn{2}{c}{Pauli Vector} \\
	    \cmidrule(lr){3-6} \\
	    & & CV & RV & CV & RV \\
	    \midrule
	    \multirow{2}{*}{FCNN} &
	    \acrshort{oa} 
	    & $79.81 \pm 7.37$ 
            & $42.88 \pm 5.75$ 
            & $\textbf{91.90} \pm 1.04$
	    & $89.86 \pm 1.14$  \\
	    & \acrshort{aa} 
	    & $61.92 \pm 4.32$  
            & $50.82 \pm 3.65$ 
	    & $\textbf{80.15} \pm 0.84$
            & $79.05 \pm 1.37$ \\
	    \midrule
	    \multirow{2}{*}{CNN} &
	    \acrshort{oa} 
	    & $89.18 \pm 0.35$  
            & $87.79 \pm 0.23$ 
	    & $\textbf{90.55} \pm 0.69$ 
            & $89.14 \pm 0.67$  \\
	    & \acrshort{aa} 
	    & $\textbf{88.40} \pm 0.30$ 
            & $87.29 \pm 0.19$  
	    & $86.97 \pm 0.91$  
            & $84.77 \pm 0.82$  \\
	    \midrule
	    \multirow{2}{*}{MLP} &
	    \acrshort{oa} 
	    & $\textbf{88.53} \pm 0.11$  
            & $87.00 \pm 0.22$ 
	    & $69.38 \pm 0.26$  
            & $70.09 \pm 0.05$   \\
	    & \acrshort{aa} 
	    & $\textbf{87.02} \pm 0.09$      
            & $86.21 \pm 0.17$  
	    & $66.89 \pm 0.20$  
            & $67.25 \pm 0.07$  \\
            \hline \hline 
            \\\\[-1.\medskipamount]
            \multicolumn{2}{c}{Loss} & \multicolumn{2}{c}{Coherency Matrix} 
            & \multicolumn{2}{c}{Pauli Vector} \\
	    \cmidrule(lr){3-6} \\
	    & & CV & RV & CV & RV \\
            \midrule
	    \multirow{2}{*}{CNN} &
	    \acrshort{oa} 
	    & $88.83 \pm 0.34$ 
            & $87.60 \pm 0.30$
	    & $\textbf{89.60} \pm 0.66$ 
            & $86.47 \pm 0.74$ \\
	    & \acrshort{aa} 
	    & $\textbf{87.89} \pm 0.32$
            & $86.83 \pm 0.28$ 
	    & $85.36 \pm 0.81$
            & $81.79 \pm 0.72$ \\
	    \midrule
	    \multirow{2}{*}{MLP} &
	    \acrshort{oa} 
	    & $\textbf{89.68} \pm 0.57$ 
            & $87.53 \pm 0.11$
	    & $65.67 \pm 0.26$
            & $67.34 \pm 0.25$ \\
	    & \acrshort{aa} 
	    & $\textbf{87.27} \pm 0.37$ 
            & $86.13 \pm 0.08$ 
	    & $63.70 \pm 0.19$ 
            & $64.71 \pm 0.22$ \\
        \bottomrule
	\end{tabular}
\end{table}

Throughout the simulations, \acrshort{rv-mlp} obtained higher results when not splitting the dataset, whereas \acrshort{cv-mlp} outperformed the real-valued model when splitting it. This means that \acrshort{rv-mlp} can learn better if the train and validation are closer to each other but cannot generalize better for less similar datasets. This might be explained by \acrshort{rv-mlp} having more tendency to overfit the data as was shown by References \cite{barrachina2020complex, ko2022coshnet}.

\section{CONCLUSIONS} \label{sec:conclusion}

We performed an exhaustive comparison of a total of three complex-valued networks against their real-valued equivalent on two input representations of \acrshort{polsar} images for segmentation applications, we show that complex models generalize better except for the \acrshort{mlp} without dataset splitting. For \acrshort{fcnn}, the Pauli vector worked better, as \acrshort{mlp} can not naturally perform local filtering operations, and it cannot deal with the speckling noise, for what the coherency matrix worked better. Finally, \acrshort{cnn} performance was not decisive on what input representation was better. 
We then performed some dataset pre-processing to reduce the correlation between the training, validation, and test datasets and repeated the results obtaining similar conclusions as before, effectively lowering the accuracy of models, which could achieve over $99\%$ accuracy to less than $94\%$.

Finally, we tried to reduce the difference between \acrshort{oa} and \acrshort{aa} scores by using two different methods.
Although both balancing methods worked correctly in reducing the gap between \acrshort{oa} and \acrshort{aa}, mainly increasing the latest, dataset balancing worked better. In \acrshort{cnn} and \acrshort{mlp} models, the wide availability of classes meant this method did not impact the performance significantly as it had the same number of training examples per class. We only tried a simple way for the weighted loss from several, so more research could be done in this area to increase performance.


\bibliographystyle{ieeetr}
\bibliography{sn-bibliography}      

\end{document}